\documentclass[11pt,letterpaper]{article}
\usepackage{cogsys}
\usepackage[T1]{fontenc}
\usepackage{times}
\usepackage{graphicx} 

\usepackage{natbib}
\cogsysheading{XI}{2024}{1-6}{6/2024}{6/2024}

\usepackage{bookmark}
\usepackage[utf8]{inputenc} 
\usepackage[english]{babel}
\usepackage{refstyle}
\usepackage{csquotes}
\usepackage{hyperref}       
\usepackage{url}            
\usepackage{booktabs}       
\usepackage{tabularx}
\usepackage{multirow}
\usepackage{amsfonts}       
\usepackage{nicefrac}       
\usepackage{microtype}      
\usepackage{xcolor}         
\usepackage{pgfplots}
\pgfplotsset{compat=1.18}
\usepackage{mathtools}

\usepackage[nounderscore]{syntax} 
\usepackage{caption}
\usepackage{subcaption}
\usepackage{mdframed}
\usepackage{amsmath}
\usepackage{amssymb}
\usepackage{graphicx}
\usepackage{listings}
\usepackage{multirow}
\usepackage{cprotect}

\usepackage{xcolor}
\usepackage{soul}
\newtoggle{BLIND}
\newtoggle{COMMENTS}

\settoggle{BLIND}{true}   
\settoggle{COMMENTS}{false} 

\newcommand\com[1]{%
    \iftoggle{COMMENTS}{%
        \protect\sethlcolor{yellow}\hl{#1} } {%
        \ignorespaces%
    }%
} 

\newcommand\blind[1]{%
    \iftoggle{BLIND}{%
        (blinded for review) }{%
    #1 } }

\ShortHeadings{Learning Explainable Representations of Complex Game-playing Strategies} {Krishnan, Potts, Jhala,
Khadilkar, Karande \& Martens}

\begin{document}

\title{Learning Explainable Representations of Complex Game-playing Strategies}

\author{Abhijeet Krishnan}{akrish13@ncsu.edu}
\author{Colin M. Potts}{cmpotts@ncsu.edu}
\author{Arnav Jhala}{ahjhala@ncsu.edu}
\address{Venture IV 420, North Carolina State University, Raleigh, NC 27606 USA}
\author{Harshad Khadilkar}{harshadk@iitb.ac.in}
\address{Department of Aerospace Engineering, Indian Institute of Technology Bombay, Powai, Mumbai 400076, Maharashtra,
    India}
\author{Shirish Karande}{shirish.karande@tcs.com}
\address{Tata Research Development and Design Centre, Pune 411013, Maharashtra, India}
\author{Chris Martens}{c.martens@northeastern.edu}
\address{Meserve 138 (Boston campus), Khoury College of Computer Sciences, College of Arts, Media and Design, Northeastern University, Boston, MA 02115 USA}
\vskip 0.2in

\begin{abstract}
    As part of learning to play complex games, human players develop develop abstractions for concepts and strategies of gameplay consistent with game rules to improve their performance. These concepts are applied to explain other players' actions, and to inform their own
    actions in-game. Understanding other players' strategies is a crucial part of such improvement, but requires time and
    effort. In this paper, we propose a strategy similar to human cognition for training RL agents to synthesize learned strategies and policies as executable procedures based on sequences of gameplay actions. We present
    methods to automatically learn such programs to play chess and to solve tasks in a grid-based environment. We show
    that the learned strategies produce effective actions, and can be learned from gameplay data.
\end{abstract}

\section{Introduction}
\label{introduction}

Researchers in varied fields agree that people create \textit{mental models} as shorthand for understanding the human
experience~\citep{Boyan_Sherry_2011}. Mental models are described as \enquote{mechanisms whereby humans are able to
    generate descriptions of system purpose and form, as explanations of system functioning and observed system states, and
    as predictions (or expectations) of future system states}~\citep{rouse1992role}. In the context of games, players refine
their mental models through solitary practice or formal instruction, discussion of strategies among peers, or
investigation of underlying game mechanics~\citep{Boyan2018}. Evidence suggests that when provided with optimal game
strategies, players are able to transfer them to their game play~\citep{paredes2002transfer}. However, acquiring good
strategies involves many hours of deliberate practice, or access to expert knowledge, which may not be easily available.
Automatically learning good game-playing strategies that players can use to refine their mental model would help
overcome this challenge.

The problem of \textit{strategy synthesis} can be defined as the automatic learning of paremeterized gameplay action sequences that can be applied as procedures in games, in some
suitably interpretable representation, that can be understood by a player to change their mental model and measurably
improve their performance in-game. Existing work focuses mostly on the computational learning algorithms and
representations of strategies. For example, \citet{butler2017synthesizing} are able to learn strategies using constraint
satisfaction as condition-action rules to solve the puzzle game of Nonograms. \citet{canaan2018evolving} use a genetic
algorithm to learn strategies represented using domain-specific rules for the card game \textit{Hanabi}. A recent line
of work models a strategy as an executable program in a domain-specific programming language (DSL), which can then be
learned by applying techniques from program
synthesis~\citep{de2018evolving,marino2021programmatic,marino2022evolving,medeiros2022weakest}. Considerably less
attention has been devoted to investigating how well the learned strategies can be understood by players, and which
factors contribute to their interpretability.

Our contributions in this work are two-fold. First, we present a cognitively-inspired strategy model for chess that is
based on existing models of chess strategy found in the game's literature. We present a learning algorithm and metrics
to evaluate the effectiveness of a strategy. We show that the learned strategies can approximate a human beginner better
than a random baseline. Second, we present a learning algorithm for programmatic representations of strategies based on
the transformer model~\citep{vaswani_attention_2017}. We show that it is competitive with prior state-of-the-art while
being much more sample efficient.

\section{Strategy Synthesis}
\label{strategy-synthesis}

We define strategy synthesis through the lens of reinforcement learning (RL). RL seeks to find an optimal policy
$\pi_{*}(\cdot, \theta)$ parametrized by a model $\theta$ that maximizes the expected return in a Markov Decision Process (MDP)
$\langle \mathcal{S}, \mathcal{A}(s), \mathcal{P}, \mathcal{R}, \gamma \rangle$. Our definitions of these terms are
based on the treatment by \citet{sutton2018reinforcement}.

Strategy synthesis attempts to solve the same problem as RL, with the strategy model being the parametrization $\theta$
of a policy, and the synthesis algorithm being the optimization procedure. Unlike RL, strategy synthesis has the
additional constraint that the learned strategy model must be \textit{aligned} with the mental model of the player.
Model matching theory posits that the \enquote{alignment of game models and external situations can facilitate the
player's transfer of mental models between games and external situations}~\citep{Boyan2018}. Alignment is framed as a
matter of creating accurate facsimiles of the game, and providing the necessary tools and scaffolds to help players
\enquote{make sense of} the strategy~\citep{Martinez-Garza_Clark_2017}. However, it is unclear how alignment may be
meaningfully evaluated.

As a substitute for alignment between a learned strategy and a player's mental model, we propose using computational
metrics for model \textit{interpretability} derived from the field of explainable AI. Interpretability is defined as
\enquote{the ability to not only extract or generate explanations for the decisions of the model, but also to present
this information in a way that is understandable by human (non-expert) users to, ultimately, enable them to predict a
model's behaviour}~\citep{puiutta2020explainable}. Some metrics that have been found to correlate with interpretability
are the number of cognitive chunks used and the model size~\citep{lage2019human}.

\section{Learning Chess Strategies}
\label{chess-strategies}

In this section, we describe our strategy model for chess inspired by the concept of \textit{chess tactics}. We present
a learning algorithm based on inductive logic programming that learns chess strategies from gameplay data. We evaluate
the learned strategies and show that our algorithm can learn strategies that approximate a human beginner better than a random baseline.

\subsection{Chess Strategy Model}
\label{sec:chess-strategy-model}

We model a chess strategy as a first-order logic rule expressed in Prolog~\citep{wielemaker2003prolog} using a
domain-specific predicate vocabulary $\mathcal{P}$. As seen in \Figref{chess:tactic-model}, the rule head of the
strategy consists of the variables \verb|Position|, \verb|From| and \verb|To|. The input state is described by
\verb|Position|, which is also expressed in first-order logic using $\mathcal{P}$. \verb|From| and \verb|To| describe
the output action, namely, the move which begins from the square \verb|From| and ends on the square \verb|To|.

\begin{figure}[ht]
    \centering
    \verb|tactic|(Position, From, To)~$\leftarrow$ \\
    \begin{tabular}{rl}
         & \verb|feature_1| ($\cdots$), \\
         & \verb|feature_2| ($\cdots$), \\
         & $\vdots$                     \\
         & \verb|feature_n| ($\cdots$)
    \end{tabular}
    \cprotect\caption[Chess Strategy Model]{Our chess strategy model expressed in Prolog pseudocode. Every
        \verb|feature_i| clause is a rule defined in the predicate vocabulary.}
    \figlabel{chess:tactic-model}
\end{figure}

Our usage of first-order logic to model chess tactics is motivated by the following reasons ---

\begin{enumerate}
    \item Chess tactics are an important concept that human players use to think about chess~\citep{szabo1984computer}
          and are useful in chess education~\citep{gobet2006training}.
    \item First-order logic has been extensively used to model chess
    patterns~\citep{berliner1975representation,pitrat1977chess,wilkins1979using,huberman1968program,bramer1977representation,bratko1982knowledge}.
    \item Logic rules are commonly acknowledged to be interpretable and have a long history of
          research~\citep{zhang2021survey}.
\end{enumerate}

To evaluate the learned strategies, we define the metrics of \textit{coverage} and \textit{divergence} for a strategy $\sigma$ on a set of positions $P$ as follows ---

\begin{align}
    \text{Coverage}(\sigma, P) & \doteq \frac{|P_\mathrm{A}|}{|P|} \eqlabel{coverage}
\end{align}

\begin{align}
     \text{Divergence}_E(\sigma(\cdot), P) \doteq \frac{1}{|P_\mathrm{A}|} \sum_{(s,a_1) \in P_\mathrm{A}} \sum_{a_2 \in \mathcal{A}(s)} \sigma(a_2|s) d_E(s, a_1, a_2) \eqlabel{divergence}
\end{align}

where $P_A$ is the set of positions in $P$ where the strategy $\sigma$ is applicable, and $d_E$ is a distance measure between actions $a_1$ and $a_2$ in state $s$.

\subsection{Learning Chess Strategies using ILP}
\label{sec:learning-chess-strategies}

Inductive logic programming (ILP) is a form of symbolic machine learning that aims to induce a hypothesis (a set of
logical rules) that generalizes given training examples.~\citep{cropper2022inductive}. An ILP problem is specified by
three sets of Horn clauses~---

\begin{itemize}
    \item $B$, the background knowledge,
    \item $E^+$, the set of positive examples of the concept, and
    \item $E^-$, the set of negative examples of the concept.
\end{itemize}

The ILP problem is to induce a hypothesis $H \in \mathcal{H}$ (an appropriately chosen hypothesis space) that, in
combination with the background knowledge, entails all the positive examples and none of the negative examples.
Formally, this can be written as~---
\begin{align*}
     & \forall e \in E^+, H \cup B \models e \text{ (i.e., }H \text{ is \textit{complete)}}       \\
     & \forall e \in E^-, H \cup B \not\models e \text{ (i.e., }H \text{ is \textit{consistent)}}
\end{align*}

We formalize the problem of learning a chess strategy as an ILP problem $\langle E^+, E^-, B \rangle$. $E^+$ is a set of
$\langle \text{position}, \text{move} \rangle$ tuples where the move made in the position is drawn from a target policy.
$E^-$ is a set of $\langle \text{position}, \text{move} \rangle$ tuples where the move made in the position is
\textit{not} drawn from the target policy. $B$ is the predicate vocabulary used to express positions, moves and chess
strategies. We learn a hypothesis $H$ that maximizes the number of examples entailed in $E^+$ and minimizes the number
of examples entailed in $E^-$. The hypothesis $H$ is our chess strategy $\sigma$.

To solve this ILP problem, we use a Prolog-based ILP system called Popper~\citep{cropper2021learning}. We modify Popper
to constrain it to produce legal moves and to prune strategies that are never applicable or whose recall on the training
set is less than an empirically-determined threshold.

\begin{figure}[ht]
    \centering
    $\begin{aligned} \texttt{fork} & \text{(Position,From,To)}~\leftarrow                         \\
                              & \texttt{make\_move}\text{(From, To, Position, NewPosition),} \\
                              & \texttt{attacks}\text{(To, Square1, NewPosition),}           \\
                              & \texttt{attacks}\text{(To, Square2, NewPosition),}           \\
                              & \texttt{different\_pos}\text{(Square1, Square2),}            \\
        \end{aligned}$ \cprotect\caption[An interpretation of the \textit{fork} tactic from the chess literature using
        our predicate vocabulary]{An interpretation of the \textit{fork} tactic from the chess literature using our
        predicate vocabulary. The first \verb|attacks| clause states that the piece at \verb|To| attacks the opposing
        piece at \verb|Square1| in the current position.}
    \figlabel{validation:tactic-fork}
\end{figure}

\subsection{Evaluation}
\label{sec:chess-evaluation}

To evaluate our system for chess strategy synthesis, we trained it using a dataset of 1297 (position, move) pairs sampled
from online games played by beginner players, and evaluated it on a held-out set of 100 pairs using the metrics of accuracy,
coverage and divergence. To serve as the oracles for calculating the distance metric, we used the chess engines Stockfish
14~\citep{stockfish14} and Maia-1600~\citep{mcilroyyoung2020maia}. As baseline strategies, we used the aforementioned
chess engines, and the random strategy. We report the results for each of accuracy, coverage and divergence as a histogram with 20
buckets. We see that when using Stockfish 14 as a distance metric, the learned strategies are better at imitating the training set
than a random baseline.

\begin{figure}[htbp]
    \centering
    \begin{subfigure}{\textwidth}
        \centering
        \includegraphics[width=0.5\textwidth]{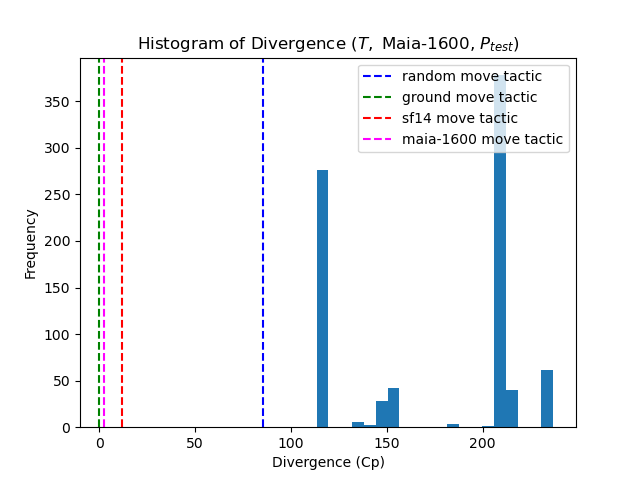}
        \caption{Divergence histogram for $T$ evaluated using Maia-1600}
        \figlabel{div-t1600-m1600}
    \end{subfigure}
    \hfill
    \begin{subfigure}{\textwidth}
        \centering
        \includegraphics[width=0.5\textwidth]{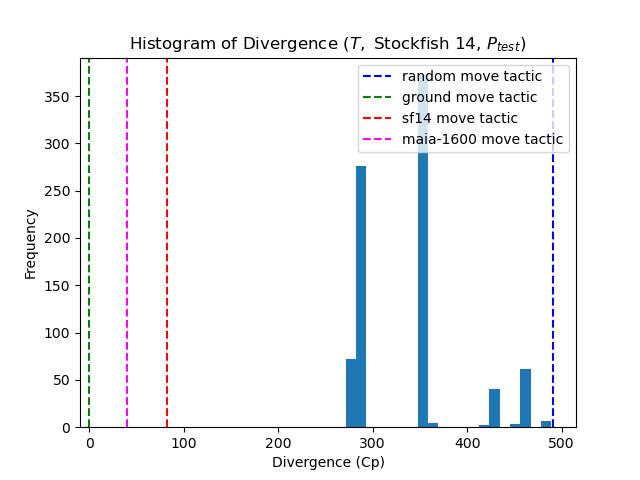}
        \caption{Divergence histogram for $T$ evaluated using Stockfish 14}
        \figlabel{div-t1600-sf14}
    \end{subfigure}
    \caption{Divergence histograms for $T$}
    \label{fig:both}
\end{figure}

\begin{figure}[htbp]
    \centering
    \begin{subfigure}{\textwidth}
        \centering
        \includegraphics[width=0.5\columnwidth]{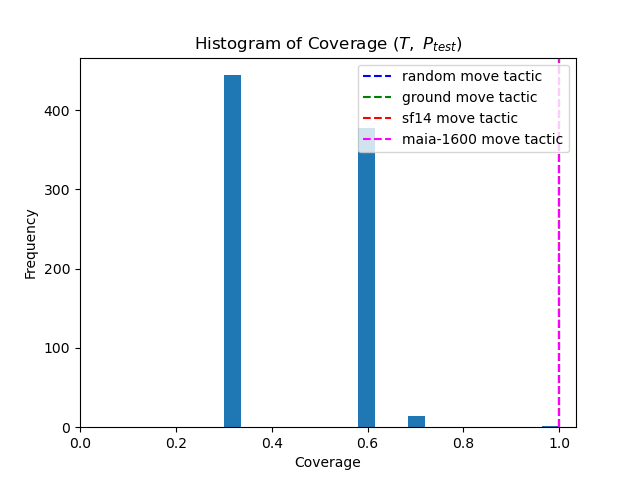}
    \caption{Coverage histogram for $T$}
    \figlabel{cov-t1600}
    \end{subfigure}
    \hfill
    \begin{subfigure}{\textwidth}
        \centering
        \includegraphics[width=0.5\columnwidth]{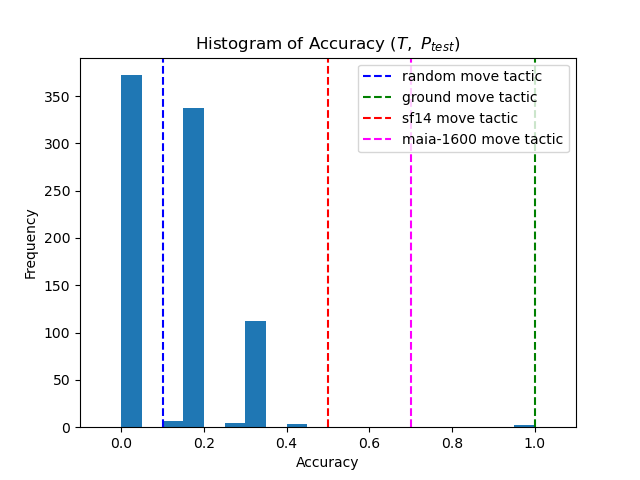}
    \caption{Accuracy histogram for $T$}
    \figlabel{acc-t1600}
    \end{subfigure}
\end{figure}

\section{Learning Programmatic Strategies with Transformers}
\label{dt}

In this section, we present a novel method for learning programmatic strategies using transformers that is competitive
with state-of-the-art methods while being more sample efficient. We first describe how problem of synthesizing
programmatic strategies can be modeled as a reinforcement learning problem. We then present our method, which applies
the decision transformer model to discrete environments. We evaluate our method using tasks in a grid-based programming
environment and show that the strategies learned by our method are competitive with a state-of-the-art programmatic
policy synthesis algorithm, while being much more sample efficient.

\subsection{Programmatic Policy Synthesis as a Reinforcement Learning Problem}
\label{sec:dt:background}

Programmatic policy synthesis is the problem of approximating a target policy using a program, as specified by some DSL.
We cast this as a RL problem by modeling the program synthesis environment as an MDP. Formally, given a DSL $G=(V,\Sigma,R,S)$~\citep{sipser1996intro} which
defines the space of possible programmatic policies, and an MDP $\mathcal{M}_{\text{exec}}$ in which a programmatic policy $\sigma$
can execute and obtain reward, we define the programmatic policy synthesis task as that of finding an optimal policy in the MDP $\mathcal{M}_{\text{syn}}$, where:

\begin{itemize}
    \item $\mathcal{S}_{\text{syn}} \doteq \{w \in (V \cup \Sigma)^*\ |\ S \xRightarrow[]{*} w\}$ and is the current sequence
          of terminals and non-terminals in the partially expanded program $(w_0, w_1, \cdots, w_i, \cdots)$.
    \item $\mathcal{A}_{\text{syn}} \doteq R \times V \times \mathbb{N}$ and is the space of all actions consisting of applying a rule $r$ to a
          non-terminal symbol $v$ located at index $i$ in the current state. \com{Better formalization?}
    \item $\mathcal{P}_{\text{syn}}$ is the transition function which represents the deterministic transition to a state
          where the production rule has been applied to the selected non-terminal symbol if such an action was valid. If
          not, the state remains unchanged. \com{how to better formalize this?}
    \item $d_{0,\text{syn}}({S}) = 1$
    \item $r_{\text{syn}} \doteq \begin{cases} r_{\text{exec}}, & \text{if current state } s \in
              \mathcal{S}_{\text{syn}} \text{ is terminal}      \\
              0,               & \text{otherwise}\end{cases}$
    \item $\gamma_{\text{syn}} = 1$ \com{Justification?}
\end{itemize}

Given this formulation, we can learn strategy models in the form of a program (in some DSL) using an RL algorithm.

\subsection{Applying Decision Transformers to Programmatic Policy Synthesis}
\label{sec:dt:method}

Decision transformers were proposed by \citet{chen_decision_2021} as an architecture to apply the sequence-modeling
capabilities of transformers~\citep{vaswani_attention_2017} to sequential decision problems. Trajectories $\tau$ are
represented as:

\begin{equation}
      \tau = \left( \widehat{R}_1, s_1, a_1, \widehat{R}_2, s_2, a_2, \cdots, \widehat{R}_T, s_T, a_T \right)
\end{equation}

where rewards $r_t$ are modeled as the returns-to-go $\widehat{R}_t = \sum_{t'=t}^{T} r_{t'}$. States, actions and
returns are embedded using separate, fully-connected layers. Additionally, an embedding is learned for each timestep and
added to each token. This is different from the standard positional embedding used by transformers, as one timestep
corresponds to three tokens. The tokens are then processed by a GPT~\citep{radford2018improving} model, which predicts
future actions via autoregressive modeling.

The original decision transformer architecture cannot handle environments with discrete actions like
$\mathcal{M}_{\text{syn}}$, and hence we make the following modifications to the architecture:
\begin{itemize}
      \item \textbf{Action Masking}: To ensure that the model does not sample invalid actions, we mask out invalid
            actions by following the approach described in \citet{huang_closer_2022}. Informally, given the unnormalized
            logits representing action predictions, we mask out invalid actions by setting their logits to $-\infty$ and
            apply a softmax to obtain a probability distribution.
      \item \textbf{Cross-entropy loss}: The original architecture used a simple L2 loss since actions were continuous.
            For discrete actions, we use a cross-entropy loss as is usual with classification
            tasks~\citep{bishop2006pattern}.
      \item \textbf{Sampling from action distribution}: The original architecture uses a regression layer to predict the
            continuous-valued action. Since we are dealing with discrete actions, we sample from the predicted action
            distribution to obtain the action.
\end{itemize}

Additionally, to better represent the sequential nature of states in the MDP, we embed the state using a sequential
model, specifically a GRU~\citep{cho-etal-2014-learning}. This follows from existing approaches to sentence embedding
for information
retrieval~\citep{Palangi_Deng_Shen_Gao_He_Chen_Song_Ward_2016,Kiros_Zhu_Salakhutdinov_Zemel_Urtasun_Torralba_Fidler_2015,Le_Mikolov_2014}.
We empirically find that this improves the model's performance when compared to embedding the state
directly~\footnote{The code release for this paper can be found at
\blind{\url{https://github.com/AbhijeetKrishnan/decision-transformer}}.}.
\com{TODO: justify using empirical data}

\com{TODO: add figure for the state-embedding layer's architecture here}

\subsection{Karel Domain}
\label{sec:dt:karel}

\begin{figure}[htbp]
    \centering
    \includegraphics[width=0.5\textwidth]{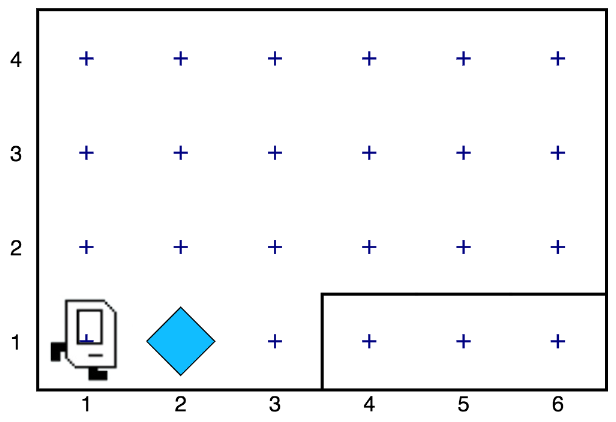}
    \caption{A sample Karel world of size $4 \times 6$. The blue diamond represents a marker. The agent cannot travel through
        walls.}
\end{figure}

Karel is a simple programming language designed for teaching programming to beginners~\citep{pattis1994karel}. Programs
written in Karel are used to control an agent that can move around a grid world and interact with markers in the world.
The agent has actions for moving and interacting with markers, and perceptions for detecting obstacles and markers.
\citet{trivedi_learning_2021} introduce six different tasks for the Karel domain with different goals and reward structures. The
initial configuration for these tasks are randomly sampled for each episode reset. They also introduce a DSL to represent programmatic
strategies for solving these tasks~(see Figure \ref{karel-dsl}).

\begin{figure}
    \centering
    \begin{mdframed}[innertopmargin=\baselineskip,innerbottommargin=0.5\baselineskip]
        \begin{grammar}
            <program> ::= `DEF' `run' `m(' <stmt> `m)'

            <stmt> ::= `REPEAT' <cste> `r(' <stmt> `r)' \alt <stmt> <stmt> \alt <action> \alt `IF' `c(' <cond> `c)' `i('<stmt>
            `i)' \alt `IFELSE' `c(' <cond> `c)' `i(' <stmt> `i)' `e(' <stmt> `e)' \alt `WHILE' `c(' <cond> `c)' `w(' <stmt>
            `w)'

            <cond> ::= `not' `c(' <cond_without_not> `c)' \alt <cond\_without\_not>

            <cond\_without\_not> ::= `frontIsClear' \alt `leftIsClear' \alt `rightIsClear' \alt `markersPresent' \alt
            `noMarkersPresent'

            <action> ::= `move' \alt `turnLeft' \alt `turnRight' \alt `putMarker' \alt `pickMarker'

            <cste> ::= `R='<int>
        \end{grammar}
    \end{mdframed}
    \caption{The domain-specific language (DSL) for constructing programs in the Karel environment.}
    \label{karel-dsl}
\end{figure}

\subsection{Evaluation}
\label{sec:dt:evaluation}

To investigate if the decision transformer can perform well on the programmatic policy synthesis task, we evaluate it
using the Karel environment and tasks. We train the decision transformer using a dataset of 50,000 programs contributed
by \citet{trivedi_learning_2021}. We use the same hyperparameters and training procedure as in
\citet{chen_decision_2021}. A learned strategy's reward is measured as its average return over 10 random initial states.
We compare the performance of strategies learned using the decision transformer with those learned by using
LEAPS~\citet{trivedi_learning_2021}. We also compare the number of candidate programs searched by each method. From the
results in Table \tabref{dt-results}, we see that the decision transformer is competitive with LEAPS in terms of
performance, while being more sample efficient.

\begin{table}[htbp]
    \centering
    \caption{Mean return ($[0, 1.1]$) of the best program found and number of unique programs explored by LEAPS and the Decision Transformer
        model on each Karel task. The standard deviation is reported in parentheses.}
    \begin{tabular}{@{}llllc@{}}
        \toprule
        \multicolumn{1}{c}{\multirow{2}{*}{\textbf{Task}}} & \multicolumn{2}{c}{\textbf{Mean Return}} & \multicolumn{2}{c}{\textbf{Unique Programs}}              \\ \cmidrule(lr){2-3}\cmidrule(lr){4-5}
        \multicolumn{1}{c}{}                               & LEAPS                                    & DT                                           & LEAPS & DT \\ \midrule
        cleanHouse                                         & 0.16 (0.13)                              & 0.23                                         & 3627  & 59 \\
        fourCorners                                        & 0.35 (0.00)                              & 0.35                                         & 9872  & 55 \\
        harvester                                          & 0.61 (0.21)                              & 0.66                                         & 11708 & 28 \\
        randomMaze                                         & 0.97 (0.04)                              & 1.0                                          & 295   & 63 \\
        stairClimber                                       & 0.74 (0.49)                              & 1.1                                          & 298   & 49 \\
        topOff                                             & 0.80 (0.11)                              & 0.66                                         & 30278 & 63 \\ \bottomrule
    \end{tabular}
    \label{tab:dt-results}
\end{table}



\section{Related Work}
\label{related-work}

\paragraph*{Neural program synthesis} Program synthesis is the task of automatically finding a program in the underlying
programming language that satisfies the user intent expressed in some form of
specification~\citep{Gulwani_Polozov_Singh_2017}. Examples of specifications include formal
rules~\citep{Manna_Waldinger_1980}, input/output pairs~\citep{neelakantan2016neural, devlin2017robustfill,
    gaunt2017differentiable, shin_improving_2018, bunel2018leveraging, gredilla_beyond_2019,chen_execution_2019,
    yang2021program}, demonstrations~\citep{sun2018neural, xu_hierarchical_2018, burke2019robotics}, natural language
instructions/prompts~\citep{liang2023code} and MDP rewards~\citep{li2020strong, trivedi_learning_2021,
    qiu2022programmatic}. Neural program synthesis focuses specifically on applying statistical learning methods based on
neural networks to the problem of program synthesis. We attempt to synthesize programs from MDP rewards using a
transformer-based approach to the task. This is motivated by the empirical success of the transformer model on sequence
learning tasks.

\paragraph*{Learning programmatic policies} A reinforcement learning policy's \textit{interpretability} refers to the
ability of a human to understand, generate explanations for and predict the policy's
behavior~\citep{puiutta2020explainable}. A result-based approach towards learning interpretable policies is by learning
a \textit{surrogate model} that approximates the original model with a simpler, ante-hoc explainable
one~\citep{speith2022taxonomies}. A line of work focuses on learning surrogate models in the form of programs (in some
programming language)~\citep{orfanos2023synthesizing,qiu2022programmatic, trivedi_learning_2021,
    inala_synthesizing_2020, verma2018pirl, verma_imitation-projected_2019}. A related line of work focuses on learning
programmatic policies for game-based MDPs~\citep{marino2021programmatic,marino2022evolving,medeiros2022weakest}. This
work investigates the use of a transformer-based architecture to the same task.

\section{Conclusion and Future Work}
\label{conclusion}

We presented the problem of strategy synthesis and explored its connections to mental model matching, model interpretability
and program synthesis. We introduced two novel approaches towards strategy synthesis -- a) a rule-based strategy model for chess
based on chess tactics, and an associated learning method, which we showed could learn strategies that better approximated a
beginner than a random baseline, and b) a transformer-based learning method for programmatic strategies that was competitive
with a state-of-the-art programmatic policy learning method while being more sample-efficient.

In future work, we plan to benchmark the decision transformer, and other programmatic policy learning methods, on a wider
range of game-based environments. We also plan to investigate the interpretability of the learned strategies, and to measure
the extent to which they are able to measurably improve a player's performance.



{\parindent -10pt\leftskip 10pt\noindent
    \bibliographystyle{cogsysapa}
    \bibliography{main}

@STRING{AN        = "Astrophys.\ Norv."}

@techreport{berliner1975representation,
  title       = {A representation and some mechanisms for a problem solving chess program},
  author      = {Berliner, Hans J},
  year        = {1975},
  institution = {Carnegie-Mellon Univ Pittsburgh PA Dept of Computer Science}
}

@book{bishop2006pattern,
  author    = {Bishop, Christopher M.},
  title     = {Pattern Recognition and Machine Learning (Information Science and Statistics)},
  year      = {2006},
  isbn      = {0387310738},
  publisher = {Springer-Verlag},
  address   = {Berlin, Heidelberg}
}

@article{Boyan_Sherry_2011,
  title        = {The Challenge in Creating Games for Education: Aligning Mental Models With Game Models},
  volume       = {5},
  rights       = {© 2011 The Authors. Child Development Perspectives © 2011 The Society for Research in Child Development},
  issn         = {1750-8606},
  doi          = {10.1111/j.1750-8606.2011.00160.x},
  abstractnote = {Abstract— Recently, there has been a great deal of interest in the use of video games for education because of the high level of engagement of players. This article blends research on game engagement from the media flow perspective with current pedagogical theory to advance 2 core design principles for a model educational game design: model matching and layering.},
  number       = {2},
  journal      = {Child Development Perspectives},
  author       = {Boyan, Andy and Sherry, John L.},
  year         = {2011},
  pages        = {82–87},
  language     = {en}
}

@article{Boyan2018,
  title    = {Model matching theory: a framework for examining the alignment between game mechanics and mental models},
  author   = {Boyan, Andy and McGloin, Rory and Wasserman, Joe A.},
  journal  = {Media and Communication},
  number   = {2},
  pages    = {126-136},
  volume   = {6},
  year     = {2018},
  issn     = {2183-2439},
  doi      = {https://doi.org/10.17645/mac.v6i2.1326},
  urn      = {https://nbn-resolving.org/urn:nbn:de:0168-ssoar-57819-8}
}

@phdthesis{bramer1977representation,
  title  = {Representation of Knowledge for Chess Endgames Towards a Self-Improving System},
  author = {Bramer, Max Arthur},
  year   = {1977},
  school = {Open University (United Kingdom)}
}

@article{bratko1982knowledge,
  title   = {Knowledge-based problem-solving in AL3},
  author  = {Bratko, Ivan},
  journal = {Machine intelligence},
  volume  = {10},
  pages   = {73--100},
  year    = {1982}
}

@inproceedings{bunel2018leveraging,
  title     = {Leveraging Grammar and Reinforcement Learning for Neural Program Synthesis},
  author    = {Rudy Bunel and Matthew Hausknecht and Jacob Devlin and Rishabh Singh and Pushmeet Kohli},
  booktitle = {International Conference on Learning Representations},
  year      = {2018},
  url       = {https://openreview.net/forum?id=H1Xw62kRZ}
}

@inproceedings{burke2019robotics,
  doi       = {10.15607/rss.2019.xv.015},
  url       = {https://doi.org/10.15607%2Frss.2019.xv.015},
  year      = 2019,
  month     = {jun},
  publisher = {Robotics: Science and Systems Foundation},
  author    = {Michael Burke and Svetlin Valentinov Penkov and Subramanian Ramamoorthy},
  title     = {From Explanation to Synthesis: Compositional Program Induction for Learning from Demonstration},
  booktitle = {Robotics: Science and Systems {XV}}
}

@inproceedings{butler2017synthesizing,
  author    = {Butler, Eric and Torlak, Emina and Popovi\'{c}, Zoran},
  title     = {Synthesizing Interpretable Strategies for Solving Puzzle Games},
  year      = {2017},
  isbn      = {9781450353199},
  publisher = {Association for Computing Machinery},
  address   = {New York, NY, USA},
  url       = {https://doi.org/10.1145/3102071.3102084},
  doi       = {10.1145/3102071.3102084},
  booktitle = {Proceedings of the 12th International Conference on the Foundations of Digital Games},
  articleno = {10},
  numpages  = {10},
  location  = {Hyannis, Massachusetts},
  series    = {FDG '17}
}

@inproceedings{canaan2018evolving,
  author    = {Canaan, Rodrigo and Shen, Haotian and Torrado, Ruben and Togelius, Julian and Nealen, Andy and Menzel, Stefan},
  booktitle = {2018 IEEE Conference on Computational Intelligence and Games (CIG)},
  title     = {Evolving Agents for the Hanabi 2018 CIG Competition},
  year      = {2018},
  volume    = {},
  number    = {},
  pages     = {1--8},
  doi       = {10.1109/CIG.2018.8490449}
}

@inproceedings{chen_decision_2021,
  title      = {Decision {Transformer}: {Reinforcement} {Learning} via {Sequence} {Modeling}},
  volume     = {34},
  shorttitle = {Decision {Transformer}},
  url        = {https://proceedings.neurips.cc/paper/2021/hash/7f489f642a0ddb10272b5c31057f0663-Abstract.html},
  urldate    = {2023-06-08},
  booktitle  = {Advances in {Neural} {Information} {Processing} {Systems}},
  publisher  = {Curran Associates, Inc.},
  author     = {Chen, Lili and Lu, Kevin and Rajeswaran, Aravind and Lee, Kimin and Grover, Aditya and Laskin, Misha and Abbeel, Pieter and Srinivas, Aravind and Mordatch, Igor},
  year       = {2021},
  pages      = {15084--15097}
}

@inproceedings{chen_execution_2019,
  title     = {Execution-Guided Neural Program Synthesis},
  author    = {Xinyun Chen and Chang Liu and Dawn Song},
  booktitle = {International Conference on Learning Representations},
  year      = {2019},
  url       = {https://openreview.net/forum?id=H1gfOiAqYm}
}

@inproceedings{cho-etal-2014-learning,
  title     = {Learning Phrase Representations using {RNN} Encoder{--}Decoder for Statistical Machine Translation},
  author    = {Cho, Kyunghyun  and
               van Merri{\"e}nboer, Bart  and
               Gulcehre, Caglar  and
               Bahdanau, Dzmitry  and
               Bougares, Fethi  and
               Schwenk, Holger  and
               Bengio, Yoshua},
  editor    = {Moschitti, Alessandro  and
               Pang, Bo  and
               Daelemans, Walter},
  booktitle = {Proceedings of the 2014 Conference on Empirical Methods in Natural Language Processing ({EMNLP})},
  month     = oct,
  year      = {2014},
  address   = {Doha, Qatar},
  publisher = {Association for Computational Linguistics},
  url       = {https://aclanthology.org/D14-1179},
  doi       = {10.3115/v1/D14-1179},
  pages     = {1724--1734}
}

@article{cropper2021learning,
  title     = {Learning programs by learning from failures},
  author    = {Cropper, Andrew and Morel, Rolf},
  journal   = {Machine Learning},
  volume    = {110},
  number    = {4},
  pages     = {801--856},
  year      = {2021},
  publisher = {Springer}
}

@article{cropper2022inductive,
  title        = {Inductive Logic Programming At 30: A New Introduction},
  volume       = {74},
  rights       = {Copyright (c)},
  issn         = {1076-9757},
  doi          = {10.1613/jair.1.13507},
  abstractnote = {Inductive logic programming (ILP) is a form of machine learning. The goal of ILP is to induce a hypothesis (a set of logical rules) that generalises training examples. As ILP turns 30, we provide a new introduction to the field. We introduce the necessary logical notation and the main learning settings; describe the building blocks of an ILP system; compare several systems on several dimensions; describe four systems (Aleph, TILDE, ASPAL, and Metagol); highlight key application areas; and, finally, summarise current limitations and directions for future research.},
  journal      = {Journal of Artificial Intelligence Research},
  author       = {Cropper, Andrew and Dumančić, Sebastijan},
  year         = {2022},
  month        = {Jun},
  pages        = {765--850},
  language     = {en}
}

@inproceedings{de2018evolving,
  author    = {de Freitas, João Marcos and de Souza, Felipe Rafael and Bernardino, Heder S.},
  booktitle = {2018 IEEE Congress on Evolutionary Computation (CEC)},
  title     = {Evolving Controllers for Mario AI Using Grammar-based Genetic Programming},
  year      = {2018},
  volume    = {},
  number    = {},
  pages     = {1-8},
  doi       = {10.1109/CEC.2018.8477698}
}

@inproceedings{devlin2017robustfill,
  title     = {{R}obust{F}ill: Neural Program Learning under Noisy {I}/{O}},
  author    = {Jacob Devlin and Jonathan Uesato and Surya Bhupatiraju and Rishabh Singh and Abdel-rahman Mohamed and Pushmeet Kohli},
  booktitle = {Proceedings of the 34th International Conference on Machine Learning},
  pages     = {990--998},
  year      = {2017},
  editor    = {Precup, Doina and Teh, Yee Whye},
  volume    = {70},
  series    = {Proceedings of Machine Learning Research},
  month     = {06--11 Aug},
  publisher = {PMLR},
  url       = {https://proceedings.mlr.press/v70/devlin17a.html}
}

@inproceedings{gaunt2017differentiable,
  title     = {Differentiable Programs with Neural Libraries},
  author    = {Alexander L. Gaunt and Marc Brockschmidt and Nate Kushman and Daniel Tarlow},
  booktitle = {Proceedings of the 34th International Conference on Machine Learning},
  pages     = {1213--1222},
  year      = {2017},
  editor    = {Precup, Doina and Teh, Yee Whye},
  volume    = {70},
  series    = {Proceedings of Machine Learning Research},
  month     = {06--11 Aug},
  publisher = {PMLR},
  url       = {https://proceedings.mlr.press/v70/gaunt17a.html}
}

@article{gobet2006training,
  title   = {Training in chess: A scientific approach},
  author  = {Gobet, Fernand and Jansen, Peter J},
  journal = {Education and chess},
  year    = {2006}
}

@article{gredilla_beyond_2019,
  title        = {Beyond imitation: Zero-shot task transfer on robots by learning concepts as cognitive programs},
  volume       = {4},
  doi          = {10.1126/scirobotics.aav3150},
  abstractnote = {Programs on a visual cognitive computer learned underlying concepts and generalized well to new settings without explicit demonstrations. Humans can infer concepts from image pairs and apply those in the physical world in a completely different setting, enabling tasks like IKEA assembly from diagrams. If robots could represent and infer high-level concepts, then it would notably improve their ability to understand our intent and to transfer tasks between different environments. To that end, we introduce a computational framework that replicates aspects of human concept learning. Concepts are represented as programs on a computer architecture consisting of a visual perception system, working memory, and action controller. The instruction set of this cognitive computer has commands for parsing a visual scene, directing gaze and attention, imagining new objects, manipulating the contents of a visual working memory, and controlling arm movement. Inferring a concept corresponds to inducing a program that can transform the input to the output. Some concepts require the use of imagination and recursion. Previously learned concepts simplify the learning of subsequent, more elaborate concepts and create a hierarchy of abstractions. We demonstrate how a robot can use these abstractions to interpret novel concepts presented to it as schematic images and then apply those concepts in very different situations. By bringing cognitive science ideas on mental imagery, perceptual symbols, embodied cognition, and deictic mechanisms into the realm of machine learning, our work brings us closer to the goal of building robots that have interpretable representations and common sense.},
  number       = {26},
  journal      = {Science Robotics},
  publisher    = {American Association for the Advancement of Science},
  author       = {Lázaro-Gredilla, Miguel and Lin, Dianhuan and Guntupalli, J. Swaroop and George, Dileep},
  year         = {2019},
  month        = {Jan},
  pages        = {eaav3150}
}

@article{Gulwani_Polozov_Singh_2017,
  title   = {Program Synthesis},
  volume  = {4},
  issn    = {2325-1107},
  doi     = {10.1561/2500000010},
  number  = {1–2},
  journal = {Foundations and Trends® in Programming Languages},
  author  = {Gulwani, Sumit and Polozov, Oleksandr and Singh, Rishabh},
  year    = {2017},
  pages   = {1–119}
}

@article{huang_closer_2022,
  title    = {A {Closer} {Look} at {Invalid} {Action} {Masking} in {Policy} {Gradient} {Algorithms}},
  volume   = {35},
  issn     = {2334-0762},
  url      = {http://arxiv.org/abs/2006.14171},
  doi      = {10.32473/flairs.v35i.130584},
  language = {en},
  urldate  = {2023-07-04},
  journal  = {The International FLAIRS Conference Proceedings},
  author   = {Huang, Shengyi and Ontañón, Santiago},
  month    = may,
  year     = {2022},
  note     = {arXiv:2006.14171 [cs, stat]}
}

@phdthesis{huberman1968program,
  title  = {A program to play chess end games},
  author = {Huberman, Barbara Jane},
  month  = {August},
  year   = {1968},
  school = {Department of Computer Science, Stanford University.}
}

@inproceedings{inala_synthesizing_2020,
  title     = {Synthesizing Programmatic Policies that Inductively Generalize},
  author    = {Jeevana Priya Inala and Osbert Bastani and Zenna Tavares and Armando Solar-Lezama},
  booktitle = {International Conference on Learning Representations},
  year      = {2020},
  url       = {https://openreview.net/forum?id=S1l8oANFDH}
}

@inproceedings{Kiros_Zhu_Salakhutdinov_Zemel_Urtasun_Torralba_Fidler_2015,
  title        = {Skip-Thought Vectors},
  volume       = {28},
  url          = {https://proceedings.neurips.cc/paper_files/paper/2015/hash/f442d33fa06832082290ad8544a8da27-Abstract.html},
  abstractnote = {We describe an approach for unsupervised learning of a generic, distributed sentence encoder. Using the continuity of text from books, we train an encoder-decoder model that tries to reconstruct the surrounding sentences of an encoded passage. Sentences that share semantic and syntactic properties are thus mapped to similar vector representations. We next introduce a simple vocabulary expansion method to encode words that were not seen as part of training, allowing us to expand our vocabulary to a million words. After training our model, we extract and evaluate our vectors with linear models on 8 tasks: semantic relatedness, paraphrase detection, image-sentence ranking, question-type classification and 4 benchmark sentiment and subjectivity datasets. The end result is an off-the-shelf encoder that can produce highly generic sentence representations that are robust and perform well in practice. We will make our encoder publicly available.},
  booktitle    = {Advances in Neural Information Processing Systems},
  publisher    = {Curran Associates, Inc.},
  author       = {Kiros, Ryan and Zhu, Yukun and Salakhutdinov, Russ R and Zemel, Richard and Urtasun, Raquel and Torralba, Antonio and Fidler, Sanja},
  year         = {2015}
}

@article{lage2019human,
  title        = {Human Evaluation of Models Built for Interpretability},
  volume       = {7},
  url          = {https://ojs.aaai.org/index.php/HCOMP/article/view/5280},
  doi          = {10.1609/hcomp.v7i1.5280},
  abstractnote = {&lt;p&gt;Recent years have seen a boom in interest in interpretable machine learning systems built on models that can be understood, at least to some degree, by domain experts. However, exactly what kinds of models are truly human-interpretable remains poorly understood. This work advances our understanding of precisely which factors make models interpretable in the context of decision sets, a specific class of logic-based model. We conduct carefully controlled human-subject experiments in two domains across three tasks based on human-simulatability through which we identify specific types of complexity that affect performance more heavily than others–trends that are consistent across tasks and domains. These results can inform the choice of regularizers during optimization to learn more interpretable models, and their consistency suggests that there may exist common design principles for interpretable machine learning systems.&lt;/p&gt;},
  number       = {1},
  journal      = {Proceedings of the AAAI Conference on Human Computation and Crowdsourcing},
  author       = {Lage, Isaac and Chen, Emily and He, Jeffrey and Narayanan, Menaka and Kim, Been and Gershman, Samuel J. and Doshi-Velez, Finale},
  year         = {2019},
  month        = {Oct.},
  pages        = {59--67}
}

@inproceedings{Le_Mikolov_2014,
  title        = {Distributed Representations of Sentences and Documents},
  issn         = {1938-7228},
  url          = {https://proceedings.mlr.press/v32/le14.html},
  abstractnote = {Many machine learning algorithms require the  input to be represented as a fixed length feature  vector. When it comes to texts, one of the most  common representations is bag-of-words. Despite their popularity, bag-of-words models have  two major weaknesses: they lose the ordering  of the words and they also ignore semantics of  the words. For example, “powerful,” “strong”  and “Paris” are equally distant. In this paper,  we propose an unsupervised algorithm that learns  vector representations of sentences and text documents. This algorithm represents each document by a dense vector which is trained to predict  words in the document. Its construction gives our  algorithm the potential to overcome the weaknesses of bag-of-words models. Empirical results show that our technique outperforms bag-of-words models as well as other techniques for  text representations. Finally, we achieve new  state-of-the-art results on several text classification and sentiment analysis tasks.},
  booktitle    = {Proceedings of the 31st International Conference on Machine Learning},
  publisher    = {PMLR},
  author       = {Le, Quoc and Mikolov, Tomas},
  year         = {2014},
  month        = jun,
  pages        = {1188–1196},
  language     = {en}
}

@misc{li2020strong,
  title         = {Strong Generalization and Efficiency in Neural Programs},
  author        = {Yujia Li and Felix Gimeno and Pushmeet Kohli and Oriol Vinyals},
  year          = {2020},
  eprint        = {2007.03629},
  archiveprefix = {arXiv},
  primaryclass  = {cs.LG}
}

@inproceedings{liang2023code,
  title     = {Code as Policies: Language Model Programs for Embodied Control},
  doi       = {10.1109/ICRA48891.2023.10160591},
  booktitle = {2023 IEEE International Conference on Robotics and Automation (ICRA)},
  author    = {Liang, Jacky and Huang, Wenlong and Xia, Fei and Xu, Peng and Hausman, Karol and Ichter, Brian and Florence, Pete and Zeng, Andy},
  year      = {2023},
  pages     = {9493–9500}
}

@article{Manna_Waldinger_1980,
  address      = {New York, NY, USA},
  title        = {A Deductive Approach to Program Synthesis},
  volume       = {2},
  issn         = {0164-0925},
  doi          = {10.1145/357084.357090},
  abstractnote = {Program synthesis is the systematic derivation of a program from a given specification. A deductive approach to program synthesis is presented for the construction of recursive programs. This approach regards program synthesis as a theorem-proving task and relies on a theorem-proving method that combines the features of transformation rules, unification, and mathematical induction within a single framework.},
  number       = {1},
  journal      = {ACM Trans. Program. Lang. Syst.},
  publisher    = {Association for Computing Machinery},
  author       = {Manna, Zohar and Waldinger, Richard},
  year         = {1980},
  month        = jan,
  pages        = {90–121}
}

@article{marino2021programmatic,
  title        = {Programmatic Strategies for Real-Time Strategy Games},
  volume       = {35},
  url          = {https://ojs.aaai.org/index.php/AAAI/article/view/16114},
  doi          = {10.1609/aaai.v35i1.16114},
  abstractnote = {Search-based systems have shown to be effective for planning in zero-sum games. However, search-based approaches have important disadvantages. First, the decisions of search algorithms are mostly non-interpretable, which is problematic in domains where predictability and trust are desired such as commercial games. Second, the computational complexity of search-based algorithms might limit their applicability, especially in contexts where resources are shared among other tasks such as graphic rendering. In this work we introduce a system for synthesizing programmatic strategies for a real-time strategy (RTS) game. In contrast with search algorithms, programmatic strategies are more amenable to explanations and tend to be efficient, once the program is synthesized. Our system uses a novel algorithm for simplifying domain-specific languages (DSLs) and a local search algorithm that synthesizes programs with self play. We performed a user study where we enlisted four professional programmers to develop programmatic strategies for mRTS, a minimalist RTS game. Our results show that the programs synthesized by our approach can outperform search algorithms and be competitive with programs written by the programmers.},
  number       = {1},
  journal      = {Proceedings of the AAAI Conference on Artificial Intelligence},
  author       = {Mariño, Julian R. H. and Moraes, Rubens O. and Oliveira, Tassiana C. and Toledo, Claudio and Lelis, Levi H. S.},
  year         = {2021},
  month        = {May},
  pages        = {381--389}
}

@article{marino2022evolving,
  title     = {Evolving interpretable strategies for zero-sum games},
  author    = {Mari{\~n}o, Julian RH and Toledo, Claudio FM},
  journal   = {Applied Soft Computing},
  volume    = {122},
  pages     = {108860},
  year      = {2022},
  publisher = {Elsevier}
}

@inproceedings{mcilroyyoung2020maia,
  author    = {McIlroy-Young, Reid and Sen, Siddhartha and Kleinberg, Jon and Anderson, Ashton},
  title     = {Aligning Superhuman AI with Human Behavior: Chess as a Model System},
  year      = {2020},
  isbn      = {9781450379984},
  publisher = {Association for Computing Machinery},
  address   = {New York, NY, USA},
  url       = {https://doi.org/10.1145/3394486.3403219},
  doi       = {10.1145/3394486.3403219},
  booktitle = {Proceedings of the 26th ACM SIGKDD International Conference on Knowledge Discovery \& Data Mining},
  pages     = {1677--1687},
  numpages  = {11},
  location  = {Virtual Event, CA, USA},
  series    = {KDD '20}
}

@article{medeiros2022weakest,
  title        = {What can we Learn Even From the Weakest? Learning Sketches for Programmatic Strategies},
  url          = {http://arxiv.org/abs/2203.11912},
  abstractnote = {In this paper we show that behavioral cloning can be used to learn effective sketches of programmatic strategies. We show that even the sketches learned by cloning the behavior of weak players can help the synthesis of programmatic strategies. This is because even weak players can provide helpful information, e.g., that a player must choose an action in their turn of the game. If behavioral cloning is not employed, the synthesizer needs to learn even the most basic information by playing the game, which can be computationally expensive. We demonstrate empirically the advantages of our sketch-learning approach with simulated annealing and UCT synthesizers. We evaluate our synthesizers in the games of Can’t Stop and MicroRTS. The sketch-based synthesizers are able to learn stronger programmatic strategies than their original counterparts. Our synthesizers generate strategies of Can’t Stop that defeat a traditional programmatic strategy for the game. They also synthesize strategies that defeat the best performing method from the latest MicroRTS competition.},
  note         = {arXiv:2203.11912 [cs]},
  number       = {arXiv:2203.11912},
  publisher    = {arXiv},
  author       = {Medeiros, Leandro C. and Aleixo, David S. and Lelis, Levi H. S.},
  year         = {2022},
  month        = {Mar},
  language     = {en}
}

@inproceedings{neelakantan2016neural,
  author    = {Arvind Neelakantan and Quoc V. Le and Ilya Sutskever},
  editor    = {Yoshua Bengio and Yann LeCun},
  title     = {Neural Programmer: Inducing Latent Programs with Gradient Descent},
  booktitle = {4th International Conference on Learning Representations, {ICLR} 2016, San Juan, Puerto Rico, May 2-4, 2016, Conference Track Proceedings},
  year      = {2016},
  url       = {http://arxiv.org/abs/1511.04834},
  bibsource = {dblp computer science bibliography, https://dblp.org}
}

@misc{orfanos2023synthesizing,
  title         = {Synthesizing Programmatic Policies with Actor-Critic Algorithms and ReLU Networks},
  author        = {Spyros Orfanos and Levi H. S. Lelis},
  year          = {2023},
  eprint        = {2308.02729},
  archiveprefix = {arXiv},
  primaryclass  = {cs.LG}
}

@article{Palangi_Deng_Shen_Gao_He_Chen_Song_Ward_2016,
  title        = {Deep Sentence Embedding Using Long Short-Term Memory Networks: Analysis and Application to Information Retrieval},
  volume       = {24},
  issn         = {2329-9304},
  doi          = {10.1109/TASLP.2016.2520371},
  abstractnote = {This paper develops a model that addresses sentence embedding, a hot topic in current natural language processing research, using recurrent neural networks (RNN) with Long Short-Term Memory (LSTM) cells. The proposed LSTM-RNN model sequentially takes each word in a sentence, extracts its information, and embeds it into a semantic vector. Due to its ability to capture long term memory, the LSTM-RNN accumulates increasingly richer information as it goes through the sentence, and when it reaches the last word, the hidden layer of the network provides a semantic representation of the whole sentence. In this paper, the LSTM-RNN is trained in a weakly supervised manner on user click-through data logged by a commercial web search engine. Visualization and analysis are performed to understand how the embedding process works. The model is found to automatically attenuate the unimportant words and detect the salient keywords in the sentence. Furthermore, these detected keywords are found to automatically activate different cells of the LSTM-RNN, where words belonging to a similar topic activate the same cell. As a semantic representation of the sentence, the embedding vector can be used in many different applications. These automatic keyword detection and topic allocation abilities enabled by the LSTM-RNN allow the network to perform document retrieval, a difficult language processing task, where the similarity between the query and documents can be measured by the distance between their corresponding sentence embedding vectors computed by the LSTM-RNN. On a web search task, the LSTM-RNN embedding is shown to significantly outperform several existing state of the art methods. We emphasize that the proposed model generates sentence embedding vectors that are specially useful for web document retrieval tasks. A comparison with a well known general sentence embedding method, the Paragraph Vector, is performed. The results show that the proposed method in this paper significantly outperforms Paragraph Vector method for web document retrieval task.},
  number       = {4},
  journal      = {IEEE/ACM Transactions on Audio, Speech, and Language Processing},
  author       = {Palangi, Hamid and Deng, Li and Shen, Yelong and Gao, Jianfeng and He, Xiaodong and Chen, Jianshu and Song, Xinying and Ward, Rabab},
  year         = {2016},
  month        = apr,
  pages        = {694–707}
}

@article{paredes2002transfer,
  title        = {Transfer of control between causal predictive judgments and instrumental responding},
  volume       = {30},
  issn         = {1532-5830},
  doi          = {10.3758/BF03192833},
  abstractnote = {Four experiments were conducted to explore outcome-specific transfer from causal predictive judgments to instrumental responding. A video game was designed in which participants had to defend Andalusia from navy and air force attacks. First, they learned the relationship between two instrumental responses (two keys on a standard keyboard) and two different outcomes (destruction of the ships or destruction of the planes). Then they learned to predict which of two different stimuli predicted which outcome. Finally, they had the opportunity of making either of the two instrumental responses in the presence of either stimulus. Transfer was shown as a preference for the response that shared an outcome with the current stimulus. The presentation of the stimulus during the test produced a decrease in the overall rate of response. Responding to a neutral stimulus in Experiments 2 and 3 suggested that this overall decrease in responding was due to a combination of the time needed to process the meaning of the stimulus and the activation of the representation of the outcome in the presence of the stimulus during the test. Transfer between predictive judgments and instrumental responding mirrors the outcome-specific Pavlovian instrumental transfer observed in conditioning studies with rats.},
  number       = {3},
  journal      = {Animal Learning \& Behavior},
  author       = {Paredes-Olay, Concepción and Abad, María J. F. and Gámez, Matías and Rosas, Juan M.},
  year         = {2002},
  month        = aug,
  pages        = {239–248},
  language     = {en}
}

@book{pattis1994karel,
  title     = {Karel the robot: a gentle introduction to the art of programming},
  author    = {Pattis, Richard E},
  year      = {1994},
  publisher = {John Wiley \& Sons}
}

@misc{stockfish14,
  author = {Romstad, Tord and Costalba, Marco and Kiiski, Joona},
  title  = {Stockfish 14},
  url    = {https://stockfishchess.org/},
  year   = {2021}
}

}


\end{document}